\documentclass[runningheads]{llncs}

\usepackage[mobile]{eccv}
\usepackage{eccvabbrv}

\usepackage{graphicx}
\usepackage{booktabs}

\usepackage[accsupp]{axessibility}  % Improves PDF readability for those with disabilities.

\usepackage[pagebackref,breaklinks,colorlinks,citecolor=eccvblue]{hyperref}
\usepackage{orcidlink}
\usepackage{xcolor}
\usepackage{colortbl}
\usepackage{multirow}
\usepackage{makecell}
\usepackage{enumitem}
\usepackage{amsmath}
\usepackage{arydshln}
\usepackage{xspace}
\usepackage{floatrow}
\newfloatcommand{capbtabbox}{table}[][\FBwidth]

\begin{document}

\title{CG-SLAM: Efficient Dense RGB-D SLAM in a Consistent Uncertainty-aware 3D Gaussian Field} 

\titlerunning{CG-SLAM}

\author{Jiarui Hu\inst{1} \and
Xianhao Chen\inst{2} \and
Boyin Feng\inst{1} \and Guanglin Li\inst{1} \and Liangjing Yang\inst{2} \and Hujun Bao\inst{1} \and Guofeng Zhang\inst{1} \and Zhaopeng Cui\inst{1}$^\dag$ }

\authorrunning{Hu et al.}

% \institute{State Key Lab of CAD\&CG, Zhejiang University\\ \email{\{hujiarui37, fengby99, zhpcui\}@gmail.com, \{liguanglin, zhangguofeng\}@zju.edu.cn, bao@cad.zju.edu.cn} \and
% ZJU-UIUC Institute, International Campus, Zhejiang University\\ \email{\{xianhao.22, liangjingyang\}@intl.zju.edu.cn}
\institute{State Key Lab of CAD\&CG, Zhejiang University\\ \and
ZJU-UIUC Institute, International Campus, Zhejiang University\\
% \and
}

\maketitle
\renewcommand{\thefootnote}{}
\footnotetext[2]{$^\dag$ Corresponding author.}

\begin{abstract}
   Recently neural radiance fields (NeRF) have been widely exploited as 3D representations for dense simultaneous localization and mapping (SLAM). Despite their notable successes in surface modeling and novel view synthesis, existing NeRF-based methods are hindered by their computationally intensive and time-consuming volume rendering pipeline. This paper presents an efficient dense RGB-D SLAM system, \ie, CG-SLAM,  based on a novel uncertainty-aware 3D Gaussian field with high consistency and geometric stability. Through an in-depth analysis of Gaussian Splatting, we propose several techniques to construct a consistent and stable 3D Gaussian field suitable for tracking and mapping. Additionally,  a novel depth uncertainty model is proposed to ensure the selection of valuable Gaussian primitives during optimization, thereby improving tracking efficiency and accuracy. Experiments on various datasets demonstrate that CG-SLAM achieves superior tracking and mapping performance with a notable tracking speed of up to 15 Hz. We will make our source code publicly available. Project page: \href{https://zju3dv.github.io/cg-slam}{https://zju3dv.github.io/cg-slam}.
  \keywords{Dense Visual SLAM \and Neural Rendering \and 3D Gaussian field}
\end{abstract}

\section{Introduction}
\label{sec:intro}
% Part 1: Dense SLAM and traditional method
Dense visual Localization and Mapping (Visual SLAM) is a long-standing problem in 3D computer vision over recent decades, which targets performing pose tracking and scene mapping simultaneously with a variety of downstream applications such as virtual/augmented reality (VR/AR), robot navigation, and autonomous driving. Traditional visual SLAM systems \cite{DVOslam} have shown accurate tracking performance across various scenes, while the underlying 3D representations (\eg, point cloud, mesh, and surfel) demonstrate limitations in facilitating highly free scene exploration, such as photorealistic scene touring, fine-grained map updating, \etc. 

% Part 2: NeRF-based method and its main issue, i.e., efficiency, FLoPs
Inspired by the Neural Radiance Field (NeRF)~\cite{nerf} in surface reconstruction and view rendering, some novel NeRF-based SLAM methods~\cite{nice-slam, point-slam} have been proposed recently and demonstrated promising performance in tracking, surface modeling, and novel view synthesis. Nevertheless, existing NeRF-based methods follow the differential volume rendering pipeline that is computation-intensive and time-consuming. Therefore they can only perform tracking and mapping by sampling a limited number of camera rays, ignoring the natural structural information in images. To avoid local optima in tracking and artifacts, they normally require many optimization steps, which makes it struggle to bring the best of both worlds concerning accuracy and efficiency.

\begin{figure*}[t]
 \centering
  \includegraphics[width=\textwidth]{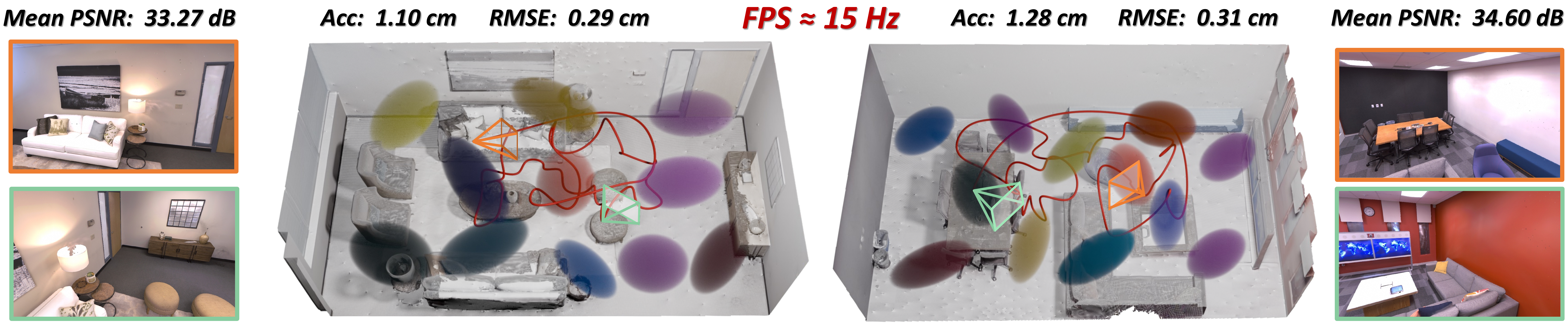}
  \caption{CG-SLAM, which adopts a well-designed 3D Gaussian field, can simultaneously achieve state-of-the-art performance in localization, reconstruction and rendering. Benefiting from 3D Gaussian representation and a new GPU-accelerated framework that is developed from a thorough derivative analysis of camera pose in 3D Gaussian Splatting~\cite{3Dgs}, CG-SLAM can perform extremely fast rendering and solve the long-standing efficiency bottleneck suffered by previous rendering-based SLAM methods. }
  \label{fig:fig1_teaser}
  \vspace{-0.5em}
\end{figure*}

% Part 3: Introduce Gaussian, and explain the challenges
Very recently the 3D Gaussian Splatting~\cite{3Dgs} method has been introduced for efficient novel view synthesis, and its rasterization-based rendering pipeline allows much faster image-level rendering, showing great potential in solving the inherent challenge of NeRF-based SLAM. However, it is nontrivial to reasonably incorporate the 3D Gaussian field in the SLAM setting. 
%First challenge
As a photorealistic view synthesis technique, the 3D Gaussian field is prone to overfitting the input images due to strong anisotropy and the lack of explicit multi-view constraints. As a result, on one hand, the 3D Gaussian splatting can not guarantee accurate modeling of 3D surfaces; on the other hand, since the Gaussians may not align with the environment's surfaces, which will lead to poor extrapolation capability and further degrade the camera tracking. 
%Second challenge
Moreover, the increase of Gaussian primitives in the mapping process will inevitably slow down the tracking efficiency.

% Part 4: Explain how CG-SLAM solve the problem.
In this paper, we introduce a real-time Gaussian splatting SLAM system, \ie, CG-SLAM, based on a novel uncertainty-aware 3D Gaussian field with high consistency and geometric stability.
To this end, we first conduct a comprehensive mathematical analysis regarding the derivatives of camera poses in the EWA (Elliptical Weighted Average) splatting process~\cite{zwicker2001ewa}, and develop a CUDA framework tailored for the SLAM task that effectively decouples the tracking and mapping components. 
%How to handle the overfitting problem
Second, in order to reduce the inherent overfitting problem, we present a scale regularization term that appropriately encourages the Gaussian ellipsoids to approximate Gaussian spheres, to reduce anisotropy and achieve a good trade-off between tracking accuracy and rendering realism.  At the same time, we observed that solely employing alpha-blending depth cannot impose effective constraints on the positions of Gaussian primitives. Thus, towards high-quality mapping, we further align median depth and alpha-blending depth to encourage Gaussian primitives to be well distributed over the scene surfaces, facilitating a consistent Gaussian field with more concentrated geometry density. 
%How to handle the second challenge.
Furthermore, in order to further improve the system's accuracy and efficiency, we design a novel depth uncertainty model to guide our Gaussian-based SLAM to focus on those stable and informative ones. 
We evaluate our system on a wide variety of RGB-D datasets, and the experimental results demonstrate that our CG-SLAM has superior performance in terms of tracking accuracy, reconstruction quality, and runtime efficiency. 

Overall, our contributions can be summarized as follows:

\begin{itemize}[leftmargin=2em]
    \item We present a new GPU-accelerated framework for real-time dense RGB-D SLAM based on a thorough theoretical analysis of camera pose derivatives in 3D Gaussian Splatting.
    \item We design multiple loss terms to build up a consistent and stable 3D Gaussian field suitable for tracking and mapping.
    \item We propose a novel depth uncertainty model, which assists our system in selecting more valuable Gaussian primitives during optimization, thereby improving tracking efficiency and accuracy.
    \item Experiments on various datasets demonstrate that our method can achieve competitive or better tracking and mapping results compared to baselines.
\end{itemize}

\section{Related Work}

\subsection{Dense Visual SLAM}
Following the seminal contributions of DTAM~\cite{DTAM} and KinectFusion~\cite{KinectFusion} to dense visual SLAM systems, there has been significant progress in developing efficient scene representation models. TSDF~\cite{KinectFusion, DTAM}, surfels~\cite{elastic, point-fusion, badslam}, voxel hashing~\cite{ChenBI13, MatthiasVoxel, voxblox}, and octrees~\cite{Franklarge, efficientoctree} have been introduced to tackle the challenges of scalability within the SLAM task. Some more advanced technologies, including bundle adjustment~\cite{badslam}, loop closure, and learning-based algorithms~\cite{tandem, cnnslam, droidslam, D3VO}, were subsequently integrated into the SLAM framework to further improve system performance. These enhancements have significantly brought better accuracy and robustness for localization and reconstruction capabilities, thus pushing the frontiers of what is achievable in dense SLAM systems. Compared with traditional methods, our CG-SLAM can reconstruct a fully dense 3D map for rich applications.

\subsection{Neural Implict Radiance Field based SLAM}
Neural radiance fields~\cite{nerf} have shown promising potential in many 3D computer vision applications, such as novel view synthesis~\cite{instant, plenoxels, mipnerf, mipnerf360}, dynamic scene modeling~\cite{tensor4d, TiNeuVox, instant_nvr, scenerflow}, and generalization~\cite{Graf, pixelnerf, pointnerf, neo360, rodin}. Recent research works~\cite{barf, imap} attempted to replace traditional maps, in tasks such as structure from motion (SFM) and SLAM, with the neural implicit field to jointly optimize scene representation and camera poses. Different kinds of neural fields brought insights into NeRF-SLAM works. NICE-SLAM~\cite{nice-slam} chose a fully covered voxel grid to store neural features, while Vox-Fusion~\cite{vox-fusion} further improved this grid to an adaptive size. Besides, Point-SLAM~\cite{point-slam} attached feature embeddings to the point cloud on object surfaces. The neural point-based method is more flexible and can encode more concentrated volume density. Co-SLAM~\cite{co-slam} adopted a hybrid representation that includes coordinate encoding and hash grids to achieve smoother reconstruction and faster convergence. In addition to the aforementioned works, some methods~\cite{nerfslam, nerfvins, orbeez} only used the neural field as a map and still performed tracking based on a traditional feature point-based visual odometry. Our CG-SLAM system can achieve better and more efficient performance in tracking, mapping, and rendering than NeRF-based methods.

\begin{figure}[t]
 \centering
  \includegraphics[width=\linewidth]{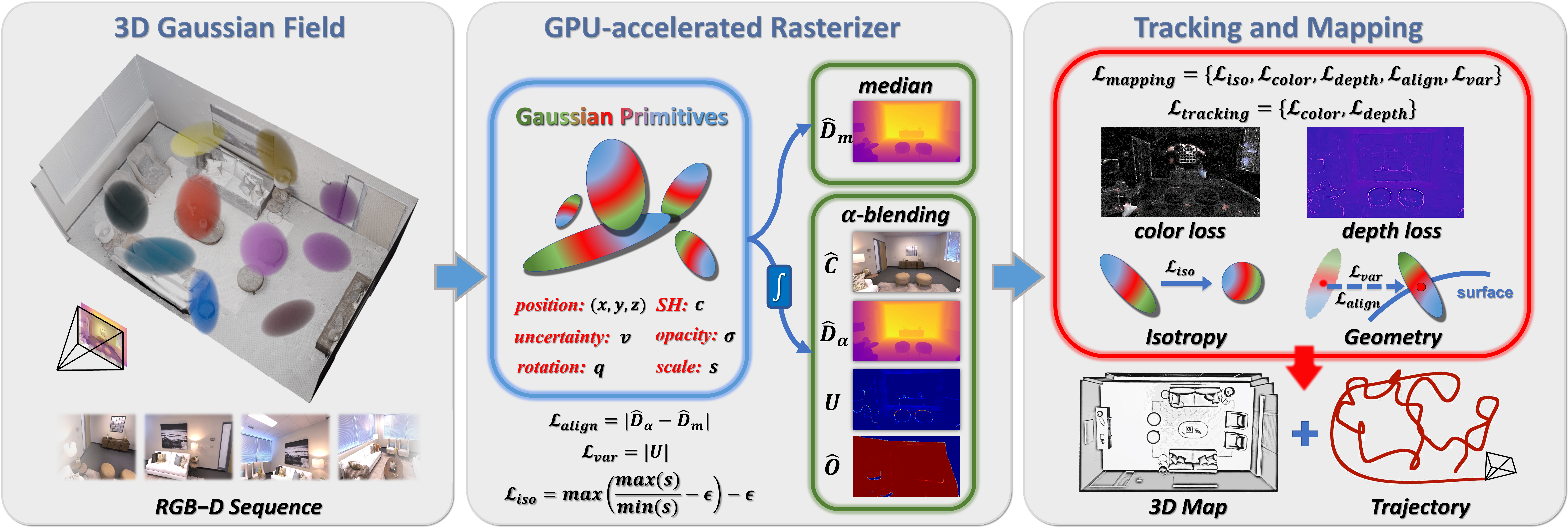}
  %\vspace{1.0em}
  \caption{\textbf{System Overview.} In a 3D Gaussian field constructed from an RGB-D sequence, we can render color, depth, opacity, and uncertainty maps through a GPU-accelerated rasterizer. Additionally, we attach a new uncertainty property to each Gaussian primitive to filter informative primitives. In the mapping process, we utilize multiple rendering results to design effective loss functions towards a consistent and stable Gaussian field. Subsequently, we employ apperance and geometry cues to perform accurate and efficient tracking.}
  \label{fig:pipeline}
\end{figure}
\subsection{3D Gaussian Splatting Field}
3D Gaussian splatting is a revolutionary novel-view synthesis approach in 3D computer vision. This approach does not contain any neural network and 
% directly attaches all properties to the Gaussian ellipsoids, which 
allows photorealistic real-time($\geq$100 FPS) rendering at 1080p resolution. 3D Gaussian splatting has influenced a wide range of research topics, such as the 3D avatar~\cite{Drivable3D, splatarmor, ash, gavatar, gaussianshell}, scene editing~\cite{GaussianEditor, GaussianEditor_Fang, gau-grouping}, image segment~\cite{featuredis}, and surface reconstruction~\cite{neusg, sugar, mdsplatting}. These studies have demonstrated its generalization and effectiveness. Similar to the neural field, some concurrent manuscripts~\cite{gsslam, splatam} have been made to reversely infer camera poses within a built 3D Gaussian field. However, these works straightforwardly apply a raw 3D Gaussian field in the SLAM framework without 
%They are short of 
specialized designs, such as anisotropy regularization and uninformative Gaussian primitive filtering. Additionally, they overlook the design on efficiency, which is the most important improvement that the Gaussian splatting technique should bring to a SLAM system.
We believe that the advantages of the 3D Gaussian field in pose optimization have not been fully explored in these works, especially in terms of efficiency, and expect to further develop an advanced Gaussian-based visual SLAM system.

% \begin{figure}[t]
%  \centering
%   \includegraphics[width=\linewidth]{fig/pipeline.pdf}
%   %\vspace{1.0em}
%   \caption{\textbf{System Overview.} In a 3D Gaussian field constructed from an RGB-D sequence, we can render color, depth, opacity, and uncertainty maps through a GPU-accelerated rasterizer. Additionally, we attach a new uncertainty property to each Gaussian primitive to filter informative primitives. In the mapping process, we utilize multiple rendering results to design effective loss functions towards a consistent and stable Gaussian field. Subsequently, we employ apperance and geometry cues to perform accurate and efficient tracking.}
%   \label{fig:pipeline}
% \end{figure}

\section{Method}
The overview of our proposed rasterization-based Gaussian SLAM system is shown in \cref{fig:pipeline}. Given a set of RGB-D sequences, our system incrementally generates a stable, consistent, and uncertainty-aware Gaussian field, serving camera pose optimization and scene geometry reconstruction. In \cref{3-1}, we briefly introduce the 3D Gaussian splatting model and the rasterization principles. We incorporated an uncertainty modeling module that utilizes the geometry prior to attach the uncertainty property on rendered images and Gaussian primitives. This strategy helps remove outliers in mapping and makes full use of informative Gaussians in tracking (\cref{3-2}). Moreover, in \cref{3-3}, we detail the Gaussian primitive management strategy and some innovative loss terms that ensure geometry stability and accuracy. Finally, by minimizing the re-rendering loss from low-uncertainty primitives, we can build a real-time and accurate tracking module (\cref{3-4}).

\subsection{Preliminary}
\label{3-1}

\textbf{Scene Representation.} 3D Gaussian Splatting defines a 3D scene as a set of anisotropic Gaussian distributions, which are associated with means $\mathbf{X}\in\mathbb{R}^3$ and covariances $\mathbf{\Sigma}\in\mathbb{R}^{3\times3}$.
% $\mathbf{X}=\{ \mathbf{X}_i|i=1,...,N, \mathbf{X}_i\in\mathbb{R}^3 \}$ and covariances $\mathbf{\Sigma} = \{ \mathbf{\Sigma}_i | i=1,...N, \mathbf{\Sigma}_i\in\mathbb{R}^{3\times3}\}$. 
To ensure that the covariance matrix remains positive semi-definite throughout the gradient descent, it is endowed with a more intuitive and comprehensible physical meaning, that is, the configuration of an ellipsoid. Specifically, $\mathbf{\Sigma}$ is simplified and decomposed into:
\begin{equation}
    \label{eq-1}
    \mathbf{\Sigma} = \mathbf{R}\mathbf{S}\mathbf{S}^T\mathbf{R}^T~,
\end{equation}
where scaling matrix $\mathbf{S} = diag([\mathbf{s}])$ is derived from the scale factor $\mathbf{s} \in \mathbb{R}^3$ and rotation matrix $\mathbf{R} \in \mathbb{R}^{3 \times 3}$ is derived from the quaternion $\mathbf{q} \in \mathbb{R}^4$. Each Gaussian ellipsoid is also assigned an opacity $\mathbf{\sigma} \in \mathbb{R}$ and spherical harmonics ($SH$) coefficients, which respectively represent the volume density and view-dependent radiance within a nearby local region.

Following Zwicker \etal~\cite{zwicker2001ewa}, given a world-to-camera rotation $\mathbf{W}$ and the Jacobian $\mathbf{J}$ of the affine approximation of the projective transformation, the EWA splatting algorithm illustrates how to approximately project a 3D Gaussian ellipsoid onto the image plane to determine its effective range and per-pixel opacity values on this image plane. We can obtain the corresponding 2D Gaussian distribution $\mathcal{N}\mathbf{(\hat{\mu}, \hat{\Sigma})}$ as:
%as in \cref{eq-2}:
\begin{equation}
    \label{eq-2}
    \mathbf{\hat{\Sigma}} = \mathbf{J}\mathbf{W}\mathbf{\Sigma}\mathbf{W}^T\mathbf{J}^T~,
\end{equation}
and $\hat{\mu}$ is the 2D pixel location of a 3D gaussian primitive center.

\textbf{Fast Rasterization-based Rendering.} Fast Gaussian splatting rasterizer enables efficient pixel-by-pixel parallel rendering, and is fully differentiable, which provides a useful GPU-accelerated framework. For an incoming frame, the rasterizer can pre-sort all visible Gaussian primitives in order of depth from near to far. In terms of color rendering, the Gaussian splatting rasterizer adopts an $\alpha$-blending solution, which accumulates radiance $c$ and opacity values $\sigma$ on a given pixel by traversing the above depth queue as follows:
\begin{equation}
    \label{eq-3}
    \hat{I} = \sum_{i=1}^N \alpha_i T_i c_i~, 
\end{equation}

\begin{equation}
    \label{eq-4}
    T_i = \prod_{k=1}^{i-1}(1-\alpha_k)~,
\end{equation}
\begin{equation}
    \label{eq-5}
    \alpha_i = \mathcal{N}\mathbf{(\hat{\mu}_i, \hat{\Sigma}_i)}  \sigma_i~, 
\end{equation}
where $\hat{I}$ is the rendered color, $T_i$ is the accumulated transmittance, $\alpha_i$ is the opacity contributed to a pixel, and $c_i$ is the color of a Gaussian primitive computed from its SH coefficients. $N$ is the number of Gaussian primitives involved in the splatting process of a pixel. In terms of depth rendering, considering the loss term designed for geometry consistency, our rasterizer provides not only $\alpha$-blending depth $\hat{D}_{alpha}$ but also the median depth $\hat{D}_{median}$:
\begin{equation}
    \label{eq-6}
    \hat{D}_{alpha} = \sum_{i=1}^N \alpha_i T_i  d_i~, 
\end{equation}

\begin{equation}
    \label{eq-7}
    \hat{D}_{median} = d_{median}~,
\end{equation}
where $d_i$ is the depth of a Gaussian primitive. For a pixel, in its splatting process, we regard a Gaussian at which the cumulative transmittance $T$ falls below $\mathbf{0.5}$ for the first time as the "median Gaussian". Its depth is recorded as $d_{median}$. $T_{median}$ is the cumulative transmittance at this median Gaussian.
\begin{equation}
    \label{eq-8}
    (T_{median} \geq 0.5)~~and~~(T_{median+1} < 0.5)~. 
\end{equation}
Besides, the accumulated opacity value $\hat{O}$ is similarly required to distinguish unobserved areas for spawning Gaussians:
\begin{equation}
    \label{eq-9}
    \hat{O} = \sum_{i=1}^N \alpha_i T_i~. 
\end{equation}

\subsection{Uncertainty Modeling}
\label{3-2}
Uncertainty model remains a trending topic in multi-view 3D reconstruction in recent decades. Inspired by ~\cite{sandstrom2023uncle}, we believe that explicitly modeling uncertainty in our 3D Gaussian field has a positive effect on increasing the ratio of informative Gaussian primitives, which is crucial for the robustness and conciseness of a SLAM system. Hence, we propose a mathematical uncertainty model suitable for RGB-D observations from two perspectives: rendering images and Gaussian primitives.

\begin{figure}[t]
  \vspace{-1em}
  \centering
  \includegraphics[width=0.6\linewidth]{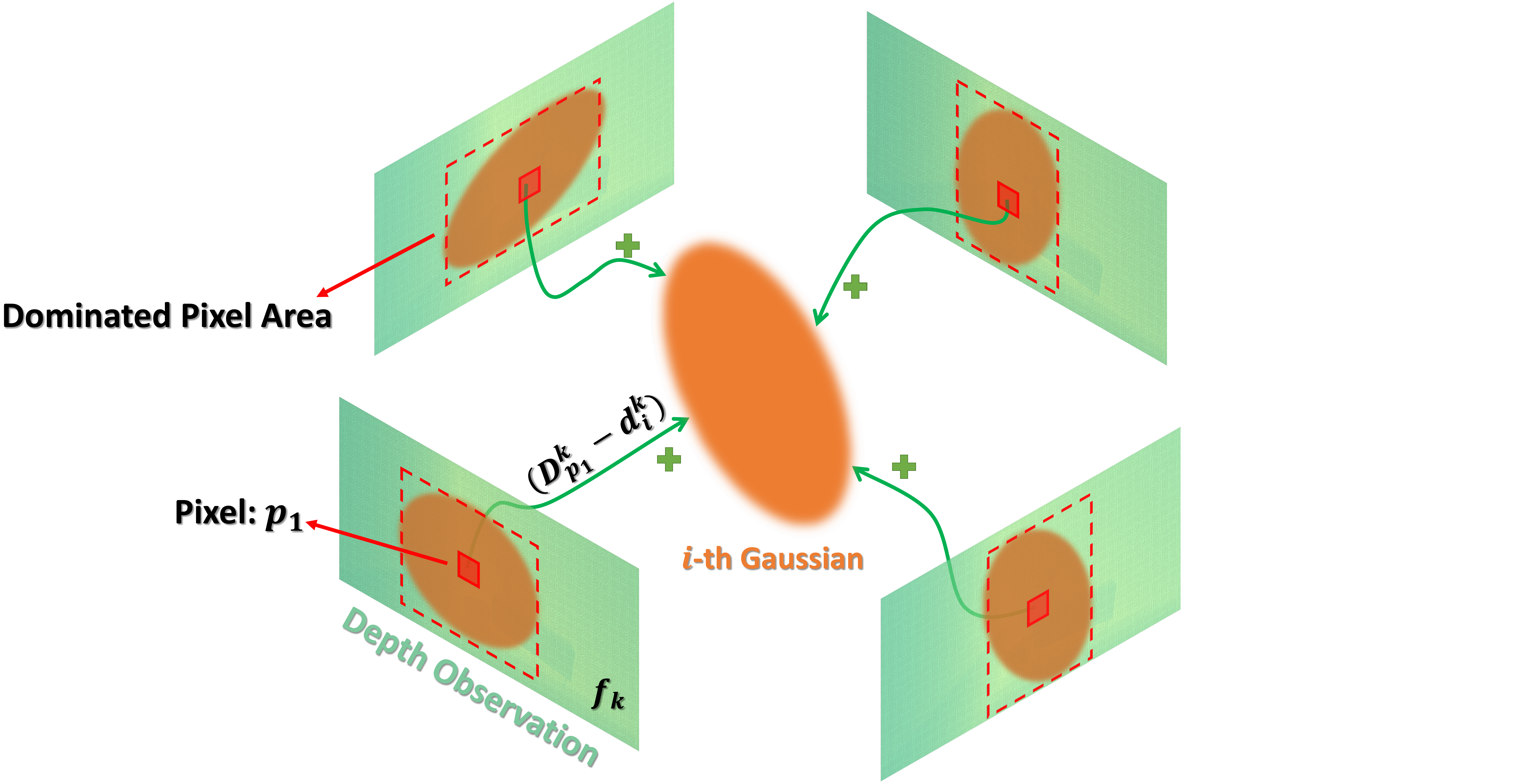}
  \caption{\textbf{Uncertainty of the Gaussian Primitives.} Uncertainty of a Gaussian primitive is derived from its dominated pixels and corresponding depth biases, reflecting the geometric value and confidence of this primitive.}
  \label{fig:uncertainty}
\end{figure}

% \vspace{-0.5em}
\textbf{Uncertainty Map.}
$\alpha$-blending depth is essentially an expected value calculated by sampling Gaussian ellipsoids along a pixel ray. Under the reasonable assumption of a normal distribution, we think that the uncertainty map is highly related to its variance. We can render an uncertainty value from the 3D Gaussian field as in \cref{eq-10}.
% \vspace{-0.5em}
\begin{equation}
    \label{eq-10}
    U = \sum_{i=1}^N \alpha_i T_i  (~d_i - D~)^2~,
    \vspace{-0.5em}
\end{equation}
where $D$ represents depth observations from the camera sensor. To mitigate drastic changes in positions of Gaussian primitives during optimization, we proposed a geometry variance loss term (\cref{eq-11}) based on the $H \times W$ uncertainty map to force them to be near the ground truth depth.
\vspace{-0.5em}
\begin{equation}
    \label{eq-11}
    \mathcal{L}_{var} =  \frac{1}{H  W} \sum_{n=1}^{HW} |U_n|~.
    \vspace{-0.5em}
\end{equation}

\textbf{Uncertainty of Gaussian primitives.}
From the perspective of geometric consistency, we design a loss term $\mathcal{L}_{align}$ as in \cref{eq-12} to align the $\alpha$-blending depth and median depth.
\vspace{-0.5em}
\begin{equation}
    \label{eq-12}
    \mathcal{L}_{align} = \frac{1}{H  W} \sum_{n=1}^{HW} |\hat{D}_{alpha}^n - \hat{D}_{median}^n|~. 
    \vspace{-0.5em}
\end{equation}
The $\alpha$-blending depth of a pixel is dominated by the Gaussian primitive with the largest weight. We call this pixel a "dominated pixel" of this maximum weight Gaussian primitive. Our alignment loss forces $\hat{D}_{alpha}$ and $\hat{D}_{median}$ to be similar, which makes this maximum weight Gaussian primitive always occur at $d_{median}$. Thus, as shown in \cref{eq-13}, the uncertainty $v_i$ of the $i$th Gaussian primitive is determined by the difference between its depth and depth observations from all its dominated pixels within a keyframe window $F =~\{f_1, f_2,...,f_k\}$. \cref{fig:uncertainty} further visualizes that a Gaussian primitive receives depth difference values from its dominated pixels in multiple viewpoints.
% Therefore, the uncertainty property of a Gaussian primitive should be regarded as a mean depth variance $v$, determined by the number of dominated pixels $M$ and corresponding depth biases. However, uncertainty extracted solely from a single viewpoint is inherently biased, so we objectively obtain this metric from co-visible frames within a keyframe window $F =~\{f_1, f_2,...,f_k\}$, as in \cref{eq-13}.
\begin{equation}
    \nu_i = \frac{1}{M_1 +...+ M_k}\sum_{f_k \in F} \sum_{p=1}^{M_k} \alpha_i^{k,p} T_i^{k,p}  (~ D_p^k - d^k_i~)^2~. 
    \label{eq-13}
\end{equation}
In a keyframe $f_k$, $\alpha_i^{k,p}$ and $T_i^{k,p}$ represent the opacity and transmittance of the $i$th Gaussian primitive on a pixel $p$. $D_p^k$ represents the depth observation on a pixel $p$ in $f_k$. $d^k_i$ is the depth value of the $i$-th Gaussian primitive at $f_k$. $\{M_1,...,M_k\}$ are the number of dominated pixels of the $i$th Gaussian primitive in different keyframes.
% In a keyframe $f_k$,~$\alpha_i^{k,p}, T_i^{k,p}, D_p^k$ represent the opacity, transmittance, and depth observation on a pixel $p$. $d^k_i$ is the depth value of the $i$-th Gaussian primitive at $k$th frame. 
Benefiting from uncertainty modeling, we can regularly detect and remove unreliable Gaussian primitives with high uncertainty exceeding a threshold $\tau$=0.025. Specifically, during the mapping optimization, primitives with $v_i>\tau$ will be manually reduced to a low-opacity level. These low-opacity Gaussian primitives can be optimized again to remove truly irreversible ones, which is a more adaptive and reasonable strategy.

\subsection{Mapping}
\label{3-3}
% For a mapping frame with the estimated pose, we combine the rendered opacity map with an empirical threshold $\psi$=0.5 to extract under-constructed pixels ($\hat{O} < \psi$) and spawn fresh primitives in the Gaussian field from these pixels.
We employ various loss functions to update Gaussian properties, aiming for a consistent and stable Gaussian field. In addition to color and $SSIM$ loss in the original 3D Gaussian splatting, previous experience from NeRF-SLAM works suggests that geometry loss is a necessary part. To overcome anisotropic interference (Arrow-shaped Gaussian primitives), we add a soft scale regularization loss in the mapping process. Note that our system performs the initialization at a slightly higher cost, \ie, more optimization iters.
\begin{equation}
    \label{eq-14}
    \mathcal{L}_{color} = \frac{1}{H  W} \sum_{n=1}^{HW} |\hat{I}_n - I_n|~, 
\end{equation}

\begin{equation}
    \label{eq-15}
    \mathcal{L}_{ssim} = SSIM(\hat{I}, I)~,
\end{equation}

\begin{equation}
    \label{eq-16}
    \mathcal{L}_{geo} = \frac{1}{H  W} \sum_{n=1}^{HW} |\hat{D}_{alpha}^n - D_n|~,
\end{equation}

\begin{equation}
    \label{eq-17}
    \mathcal{L}_{iso} = \frac{1}{G} \sum_{i \in G}max(\frac{max(\{s_i^x, s_i^y, s_i^z \})}{min(\{s_i^x, s_i^y, s_i^z \})}, \epsilon) - \epsilon~,
\end{equation}
where $I_n$ and $D_n$  represent ground-truth color and depth, $\epsilon$=1.0 is a hyperparameter that controls the level of anisotropy, and $G$ represents all visible Gaussians under the current view. 
% Benefiting from uncertainty modeling, we can regularly detect and remove unreliable Gaussian primitives with high uncertainty exceeding a threshold $\tau$=0.025. Specifically, during the mapping optimization, primitives with $v>\tau$ will be manually reduced to a low-opacity level. Finally, we can combine the global mapping loss with different weights $\omega=\{ \omega_1, ..., \omega_6\}$.
\begin{equation}
\label{eq-18}
\begin{aligned}
    \mathcal{L}_{mapping} =~ &\omega_1 \mathcal{L}_{color} + \omega_2 \mathcal{L}_{ssim} + \omega_3 \mathcal{L}_{geo}\\
    &+ \omega_4 \mathcal{L}_{align} + \omega_5 \mathcal{L}_{iso} + \omega_6 \mathcal{L}_{var}~.
\end{aligned}
\end{equation}
These loss functions customized for the SLAM task facilitate faster convergence in mapping and lay a solid foundation for subsequent tracking.

\textbf{Gaussian Management.} In initialization, we densely project Gaussian primitives into 3D space based on depth observations of the first frame. In subsequent mapping, we set an empirical threshold $\psi=0.5$ to extract unobserved or under-constructed pixels where  $\hat{O}<\psi$. Then, we utilize color and depth information on these pixels to spawn fresh Gaussian primitives. In addition, we inherited the original splitting, cloning, and removing strategy for Gaussian densification.
\subsection{Tracking}
\label{3-4}
In our system, we have proposed the first comprehensive mathematical theory on derivatives w.r.t. pose in 3D Gaussian splatting framework (refer to supplementary), and empirically discovered that the Lie algebraic representation is more advantageous for camera tracking, especially for the rotation, in a Gaussian field. The camera pose optimization, \ie, rotation and translation $\{\mathfrak{so}(3)|T \}$, mainly includes two parts: sequential tracking and sliding bundle adjustment.

\textbf{Sequential Tracking.} Given the fixed scene representation, the camera pose is initially guessed via the constant speed assumption where the last pose is transformed by the last relative transformation, and then we refine this rough pose using similar photometric and geometric losses weighted by $\lambda = \{ \lambda_1, \lambda_2 \}$.
\begin{equation}
    \label{eq-19}
    \mathcal{L}_{tracking} = \lambda_1 \mathcal{L}_{color} + \lambda_2 \mathcal{L}_{geo}~,
\end{equation}

\begin{equation}
    \label{eq-20}
    \{\mathfrak{so}(3)|T \} = \mathop{argmin}\limits_{\{\mathfrak{so}(3)|T \}}(\mathcal{L}_{tracking})~.
\end{equation}

\textbf{Sliding Bundle Adjustment.}
Cumulative error is a typical problem in SLAM, also in Gaussian-based SLAM systems. To ease it, we set up a sliding window $F$ containing $k$ co-visible keyframes and jointly optimize camera extrinsics and scene representation in this window. Due to the efficiency, we encode keyframes into a descriptor pool with a pre-trained NetVLAD~\cite{arandjelovic2016netvlad} model to determine co-visibility through the cosine similarity scores, instead of the view frustum overlap method in previous works. In addition to keyframes from NetVLAD~\cite{arandjelovic2016netvlad}, we also added the current frame and the most recent keyframes in the sliding window considering temporal associations. We still employ $\mathcal{L}_{mapping}$ in sliding bundle adjustment, where $\Psi$ is a set of all optimizable Gaussian properties.

\begin{equation}
    \label{eq-21}
    \Psi, \{\mathfrak{so}(3)|T \} = \mathop{argmin}\limits_{\Psi, \{\mathfrak{so}(3)|T \}}(\mathcal{L}_{mapping})~.
\end{equation}

\colorlet{colorFst}{green!25}
\colorlet{colorSnd}{lime!45}
\colorlet{colorTrd}{yellow!30}
\newcommand{\fs}{\cellcolor{colorFst}\textbf}
\newcommand{\nd}{\cellcolor{colorSnd}}
\newcommand{\rd}{\cellcolor{colorTrd}}

\section{Experiments}
In this section, we describe our experimental setup and validate that the proposed system can achieve improvement in both accuracy (\cref{4-1} and \cref{4-2}) and efficiency (\cref{4-3}). 
%In terms of rendering quality, we demonstrate our advantages in this aspect (Sec.~\ref{4-4}). 
We also confirmed the effectiveness of our design choices (\cref{4-5}). Additionally, we demonstrate our advantages in image rendering and capability for online third-person view rendering in supplementary. We color each cell as \colorbox{colorFst}{\bf best}, \colorbox{colorSnd}{\bf second best}, and \colorbox{colorTrd}{\bf third best}.
% We also demonstrate our advantage in image rendering and capability for online third-person view rendering (\cref{4-4}). 
% Finally, we confirmed our design choices (\cref{4-5}).
% \subsection{Experimental Setup}

\textbf{Datasets.}
To evaluate our system in various scenarios, we use three standard benchmarks: Replica~\cite{straub2019replica}, TUM~\cite{sturm2012benchmark}, and ScanNet~\cite{dai2017scannet}. The Replica dataset contains 8 available synthetic RGB-D sequences generated by Sucar \etal~\cite{imap}. We examined the generalization of our method on real-world TUM~\cite{sturm2012benchmark} and ScanNet~\cite{dai2017scannet} datasets, which contain 5 and 6 challenging scenes respectively. 

\textbf{Implementation Details.}
We run our system on a desktop equipped with an Intel i9-14900K and an NVIDIA RTX 4090 GPU. We set the learning rate of $\{\mathfrak{so}(3)|T \}$ to $\{0.0015, 0.00215\}$ in sequential tracking in all experiments. For the Replica~\cite{straub2019replica} dataset, we perform 60-iteration mapping with weights $\omega=\{$ 0.7, 0.1, 0.25, 0.25, 0.1, 0.15 $\}$  in a sliding window with $k=4$ keyframes and 15-iteration sequential pose optimization weighted by $\lambda =\{0.2,1.0 \}$. We select keyframes at an interval of 30. 
For TUM~\cite{sturm2012benchmark} and ScanNet~\cite{dai2017scannet} datasets, we use $\omega=\{$ 1.0, 0.1, 0.8, 0.5, 0.1, 0.5 $\}$, $k=4$, $\lambda=\{1.0,0.6 \}$. Faced with challenging real-world scenes, we need to extract more keyframes at an interval of 15, perform mapping at 40$\sim$50 iterations, and increase tracking iterations to 25. For further implementation details, please refer to our supplementary.

\textbf{Metrics.}
We quantitatively evaluate reconstruction quality using different 3D metrics. Given 3D triangle meshes, we compute mapping $Accuracy$ [cm], $Completion$ [cm], and $Completion~Ratio$ [<5cm \%]. Following NICE-SLAM~\cite{nice-slam}, we discard unobserved regions that are not in any viewpoints. As for tracking performance, we measure ATE RMSE~\cite{sturm2012benchmark} for estimated trajectories. 

\textbf{Baselines.}
We primarily consider state-of-the-art NeRF-SLAM works, including NICE-SLAM~\cite{nice-slam}, Co-SLAM~\cite{co-slam}, Point-SLAM~\cite{point-slam}, and Vox-Fusion~\cite{vox-fusion}, as baselines. For a fair comparison, we reproduced all results from these baselines and reported their reconstruction performance with the same evaluation mechanism. %For concurrent Gaussian-based SLAM works, such as GS-SLAM~\cite{gsslam} and SplaTAM~\cite{splatam}, we refer to the reported results in their papers.
We also add some concurrent manuscripts such as GS-SLAM~\cite{gsslam} and SplaTAM~\cite{splatam} for reference, and we directly report the results in their papers.

% %Table setting
% \colorlet{colorFst}{green!25}
% \colorlet{colorSnd}{lime!45}
% \colorlet{colorTrd}{yellow!30}
% \newcommand{\fs}{\cellcolor{colorFst}\textbf}
% \newcommand{\nd}{\cellcolor{colorSnd}}
% \newcommand{\rd}{\cellcolor{colorTrd}}

% Replica Tracking Table
\begin{table}[!t]
  \caption{\textbf{Tracking Results on the Replica Dataset~\cite{straub2019replica} (ATE RMSE [cm] $\downarrow$).} Our system consistently achieved the best performance in this dataset, both for 8 individual scenes and for the average. It is worth noting that GS-SLAM~\cite{gsslam} and SplaTAM~\cite{splatam} are concurrent with ours.}
  \vspace{-0.8em}
  \centering
  \footnotesize
  \setlength{\tabcolsep}{3pt}
  \renewcommand{\arraystretch}{1.0}
  \resizebox{0.7\linewidth}{!}{
  \begin{tabular}{@{\hspace*{1em}}lccccccccc}
    \toprule{Method}
    & \multicolumn{1}{c}{\makecell{\tt{rm-0}}} & \multicolumn{1}{c}{\makecell{\tt{rm-1}}} &  \multicolumn{1}{c}{\makecell{\tt{rm-2}}} & \multicolumn{1}{c}{\makecell{\tt{off-0}}} & \multicolumn{1}{c}{\makecell{\tt{off-1}}} & \multicolumn{1}{c}{\makecell{\tt{off-2}}}& \multicolumn{1}{c}{\makecell{\tt{off-3}}} & \multicolumn{1}{c}{\makecell{\tt{off-4}}} & Avg. \\
    \midrule
    NICE-SLAM   & 0.97      & 1.31      & 1.07      & 0.88      & 1.00      & 1.06      & 1.10      & 1.13      & 1.06    \\
    Co-SLAM       & 0.77      & 1.04      & 1.09      & 0.58      & 0.53      & 2.05      & 1.49      & 0.84      & 0.99 \\
    Point-SLAM & 0.56      & \rd{0.47} & \rd{0.30} & \rd{0.35} & 0.62      & 0.55 & 0.72      & 0.73      & 0.54 \\
    Vox-Fusion & \rd{0.40} & 0.54      & 0.54      & 0.50      & 0.46      & 0.75      & 0.50      & \rd{0.60} & 0.54    \\
    GS-SLAM        & 0.48      & 0.53      & 0.33      & 0.52      & 0.41 & 0.59      & \rd{0.46} & 0.70      & 0.50    \\
    SplaTAM       & \nd{0.31} & \nd{0.40} & \nd{0.29} & 0.47 & \nd{0.27} & \nd{0.29} & \nd{0.32} & \nd{0.55} & \nd{0.36}    \\
    \midrule
    \textbf{Ours}                & \fs{0.29} & \fs{0.27} & \fs{0.25} & \nd{0.33} & \fs{0.14} & \fs{0.28} & \fs{0.31} & \fs{0.29} & \fs{0.27} \\
    \textbf{Ours-light}   & 0.44 & \nd{0.40} & 0.34 & \fs{0.31} & \rd{0.30} & \rd{0.43} & 0.48 & 0.63 & \rd{0.42} \\
    \bottomrule
  \end{tabular}}
  \label{tab:Replica_tracking}
\end{table}

% TUM-RGBD Tracking Table
\begin{table}[t]
  \caption{\textbf{Tracking Results on the TUM-RGBD Dataset~\cite{sturm2012benchmark} (ATE RMSE [cm] $\downarrow$).} Our system achieves better tracking accuracy and lower variance in different scenarios. "-" indicates unavailable results because the related work is not open source.}
  \vspace{-0.8em}
  \centering
  \footnotesize
  \setlength{\tabcolsep}{3pt}
  \renewcommand{\arraystretch}{1.0}
  \resizebox{0.7\linewidth}{!}{
  \begin{tabular}{@{\hspace*{1em}}lcccccc}
    \toprule{Method}
    &  \tt{fr1/desk} &  \tt{fr1/desk2} &  \tt{fr1/room} & \tt{fr2/xyz} &  \tt{fr3/office} & \tt{Avg.}\\
    \midrule
    NICE-SLAM  & 4.26  & 4.99 & 34.49      & 31.73     & 3.87      & 15.87     \\
    Co-SLAM    & \nd{2.7}  & \nd{4.57} & 30.16      & 1.9       & \nd{2.6}  & 8.38 \\    
    Point-SLAM  & 4.34      & \fs{4.54} & 30.92      & 1.31      & \rd{3.48} & 8.92      \\
    Vox-Fusion  & 3.52      & 6.00      & 19.53 & 1.49      & 26.01     & 11.31     \\
    GS-SLAM     & 3.3  & -         & -          & 1.3  & 6.6       & -         \\
    SplaTAM     & 3.35      & 6.54      & \rd{11.13} & \nd{1.24} & 5.16      & \rd{5.48} \\
    \midrule
    \textbf{Ours}               & \fs{2.43} & \fs{4.54} & \fs{9.39}  & \fs{1.20} & \fs{2.45} & \fs{4.0}  \\
    \textbf{Ours-light}     & \rd{3.14} & \rd{4.73} & \nd{10.67}  & \rd{1.28} & \nd{2.60} & \nd{4.48}  \\
    \bottomrule
  \end{tabular}}
  \vspace{-0.6em}
  \label{tab:TUM-RGBD_tracking}
\end{table}
\vspace{-1em}

% ScanNet Tracking Table
\begin{table}[t]
  \caption{\textbf{Tracking Results on the ScanNet Dataset~\cite{dai2017scannet} (ATE RMSE [cm] $\downarrow$).} Our method achieves state-of-the-art tracking results in 6 scenes, and exceeded other methods on average. "-" indicates failure results in Vox-Fusion~\cite{vox-fusion}.}
  \vspace{-0.8em}
  \centering
  \footnotesize
  \setlength{\tabcolsep}{3pt}
  \renewcommand{\arraystretch}{1.0}
  \resizebox{0.7\linewidth}{!}{
  \begin{tabular}{@{\hspace*{1em}}lccccccc}
    \toprule{Method}
    &  \tt{Sc.0000} &  \tt{Sc.0059} &  \tt{Sc.0106} & \tt{Sc.0169} &  \tt{Sc.0181} & \tt{Sc.0207} & \tt{Avg.}\\
    \midrule
    NICE-SLAM   & 12.00      & 14.00      & \fs{7.90} & 10.90      & 13.40      & \rd{6.20} & 10.70 \\
    Co-SLAM      & \rd{7.18}  & 12.29      & 10.9 & \fs{6.62}  & 13.43      & 7.13      & \rd{9.37}  \\    
    Point-SLAM & 10.24      & \nd{8.29}  & 11.86     & 22.16      & 14.77      & 9.54      & 12.19      \\
    Vox-Fusion & 8.39  & 8.95  & -         & 9.50  & \rd{12.20} & 6.43 & -          \\
    SplaTAM  & 12.83      & 10.10      & 17.72     & 12.08      & \fs{11.10} & 7.46      & 11.88      \\
    \midrule
    \textbf{Ours}                & \nd{7.09}  & \fs{7.46}  & \rd{8.88} & \nd{8.16}  & \nd{11.60} & \fs{5.34} & \fs{8.08}  \\
    \textbf{Ours-light}                & \fs{6.90}  & \rd{8.36}  & \nd{8.72} & \rd{8.21}  & 12.72 & \nd{5.70} & \nd{8.44}  \\
    \bottomrule
  \end{tabular}}
  \vspace{-1.2em}
  \label{tab:ScanNet_tracking}
\end{table}

\subsection{Localization Evaluation}
\label{4-1}
We report the localization accuracy of our system in 8 Replica~\cite{straub2019replica} scenes in \cref{tab:Replica_tracking}, where we surpass all other methods by a notable margin around 25\%$\sim$75\%. In our Gaussian-based system, image-level pose optimization and the well-designed Gaussian field promote fast and stable convergence to an optimal solution. This is the reason why we have lower variances and higher accuracy. As shown in \cref{tab:TUM-RGBD_tracking}, despite noisy and sparse depth information in the real-world TUM-RGBD dataset~\cite{sturm2012benchmark}, our method still achieves better or competitive performance in 5 selected scenarios. We also benchmark our method and baselines on the similarly challenging ScanNet~\cite{dai2017scannet} to compare their tracking performance in \cref{tab:ScanNet_tracking}. Sensor data from multiple large-scale ScanNet scenes suffers from severe motion blur and specular reflections. Our method further demonstrates its effectiveness and superiority on this dataset,  excelling or maintaining competitiveness in various scenes. Extensive experiments showcase the remarkable ability of our proposed system to track and handle complex situations.

\subsection{Reconstruction Evaluation}
\label{4-2}
\begin{figure}[t]
 \centering
  \includegraphics[width=\linewidth]{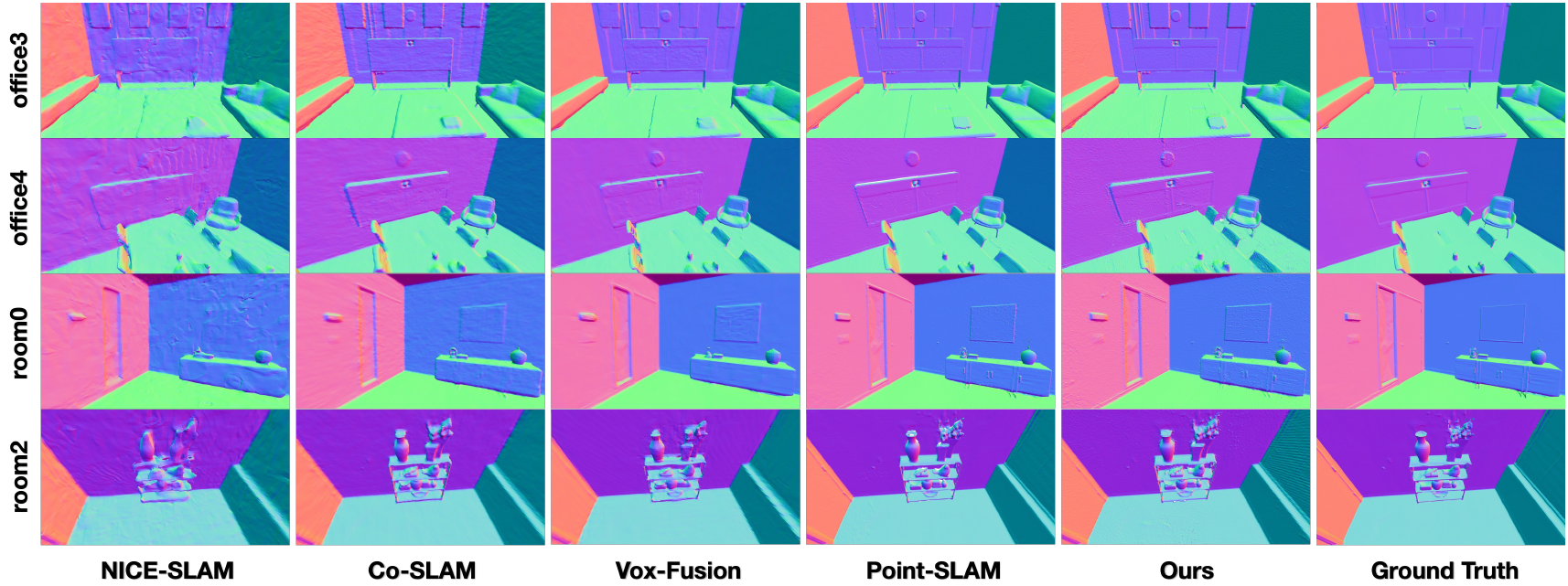}
  \caption{\textbf{Reconstruction Performance on Replica\cite{straub2019replica} Dataset.} We qualitatively compared the mesh reconstruction results from CG-SLAM and other baselines, where CG-SLAM can produce more detailed geometry at a lower computation cost.}
  \vspace{-0.6em}
  \label{fig:reconstruction}
\end{figure}

In \cref{tab:replica_rec}, we quantitatively measure the mapping performance of our proposed system, in comparison to NICE-SLAM~\cite{nice-slam}, Co-SLAM~\cite{co-slam}, Point-SLAM~\cite{point-slam}, and Vox-Fusion~\cite{vox-fusion}. It can be observed that our method outperforms all baselines on mapping accuracy. We use TSDF-Fusion~\cite{tsdf-fusion} to construct our mesh map. We achieve state-of-the-art reconstruction in observed areas. It is worth noting that the Gaussian-based method neither has a global MLP nor a fully covered feature grid, as in Co-SLAM~\cite{co-slam}. Consequently, such a system exhibits a slightly weaker hole-filling ability compared to the NeRF-based method, which explains why our system is slightly worse in the $Completion$ metric. As shown in \cref{fig:reconstruction}, we qualitatively present the ground truth mesh and mesh reconstructions from ours and other baselines. Evidently, our system achieves more detailed geometry and less noisy outliers. 
% We attribute this high-fidelity map to unique designs in the geometry consistency of the 3D Gaussian field.

\vspace{-1em}

\begin{table}[!t]
  \caption{\textbf{Reconstruction Results on the Replica~\cite{straub2019replica} Dataset.} In terms of mapping accuracy, our method can outperform all existing methods. Due to the inherent limitation of 3D Gaussian representation, our method is slightly worse in completion.}
  \vspace{-0.8em}
  \centering
  \footnotesize
  \setlength{\tabcolsep}{3pt}
  \renewcommand{\arraystretch}{1.0}
  \resizebox{0.94\linewidth}{!}{
  \begin{tabular}{@{\hspace*{1em}}llcccccccccccccccccc}
    \toprule{Method}
    & Metric & \multicolumn{1}{c}{\makecell{\tt{rm-0}}} & \multicolumn{1}{c}{\makecell{\tt{rm-1}}} &  \multicolumn{1}{c}{\makecell{\tt{rm-2}}} & \multicolumn{1}{c}{\makecell{\tt{off-0}}} & \multicolumn{1}{c}{\makecell{\tt{off-1}}} & \multicolumn{1}{c}{\makecell{\tt{off-2}}}& \multicolumn{1}{c}{\makecell{\tt{off-3}}} & \multicolumn{1}{c}{\makecell{\tt{off-4}}} & Avg. \\
    \midrule    
    \multirow{3}{*}{NICE-SLAM}
    & Acc.[cm]$\downarrow$               & 3.53 & 3.60 & 3.03 & 5.56 & 3.35 & 4.71 & 3.84 & 3.35 & 3.87 \\
    & Comp.[cm]$\downarrow$              & 3.40 & 3.62 & 3.27 & 4.55 & 4.03 & 3.94 & 3.99 & 4.15 & 3.87 \\
    & Comp.Ratio[$<\!5$cm\,\%]$\uparrow$ & 86.05 & 80.75 & 87.23 & 79.34 & 82.13 & 80.35 & 80.55 & 82.88 & 82.41 \\
    \midrule
    \multirow{3}{*}{Co-SLAM}
    & Acc.[cm]$\downarrow$               & 2.11  & 1.68  & 1.99  & 1.57  & 1.31  & 2.84  & 3.06  & 2.23  & 2.10 \\
    & Comp.[cm]$\downarrow$              & \fs{2.02}  & \fs{1.81}  & \fs{1.96}  & \nd{1.56}  & \fs{1.59}  & \fs{2.43}  & \fs{2.72}  & \fs{2.52}  & \fs{2.08} \\
    & Comp.Ratio[$<\!5$cm\,\%]$\uparrow$ & \fs{95.26} & \fs{95.19} & \fs{93.58} & \fs{96.09} & \fs{94.65} & \fs{91.63} & \fs{90.72} & \fs{90.44} & \fs{93.44} \\
    \midrule
    \multirow{3}{*}{Point-SLAM}
    & Acc.[cm]$\downarrow$               & \rd{1.45}  & \rd{1.14}  & \rd{1.19}  & \rd{1.05}  & \rd{0.86}  & \rd{1.31}  & \rd{1.57}  & \rd{1.51}  & \rd{1.26} \\
    & Comp.[cm]$\downarrow$              & 3.46  & 3.02  & \rd{2.65}  & \rd{1.65}  & 2.21  & 3.62  & 3.47  & 3.90  & 3.00 \\
    & Comp.Ratio[$<\!5$cm\,\%]$\uparrow$ & \rd{88.48} & 89.44 & \nd{90.13} & 93.39 & \rd{90.51} & \rd{86.17} & \rd{86.00} & \rd{85.74} & \rd{88.73} \\
    \midrule
    \multirow{3}{*}{Vox-Fusion}
    & Acc.[cm]$\downarrow$               & 1.77  & 1.51  & 2.23 & 1.63 & 1.44 & 2.09 & 2.33 & 2.02 & 1.88 \\
    & Comp.[cm]$\downarrow$              & \nd{2.69}  & \nd{2.31}  & \nd{2.58} & 1.87 & \nd{1.66} & \nd{3.03} & \nd{2.81} & \nd{3.51} & \nd{2.56} \\
    & Comp.Ratio[$<\!5$cm\,\%]$\uparrow$ & \nd{92.03} & \nd{92.47} & \nd{90.13} & \nd{93.86} & \nd{94.40} & \nd{88.94} & \nd{89.10} & \nd{86.53} & \nd{90.93} \\
    \midrule    
    \multirow{3}{*}{\textbf{Ours}}
    & Acc.[cm]$\downarrow$               & \fs{1.10}  & \fs{0.97}  & \fs{0.96}  & \fs{0.85}  & \fs{0.67}  & \fs{1.10}  & \fs{1.28}  & \fs{1.16}  & \fs{1.01} \\
    & Comp.[cm]$\downarrow$              & \rd{3.26}  & 2.77  & 2.79  & \fs{1.49}  & \rd{2.15}  & \rd{3.34}  & \rd{3.23}  & \rd{3.66}  & \rd{2.84} \\
    & Comp.Ratio[$<\!5$cm\,\%]$\uparrow$ & 88.26 & \rd{89.48} & \rd{89.10} & \rd{93.60} & 90.14 & 86.04 & 85.78 & 85.66 & 88.51 \\
    \midrule    
    \multirow{3}{*}{\textbf{Ours-light}}
    & Acc.[cm]$\downarrow$  & \nd{1.17} & \nd{0.98} & \nd{0.99} & \nd{0.87} & \nd{0.71} & \nd{1.2} & \nd{1.36} & \nd{1.26} & \nd{1.06} \\
    & Comp.[cm]$\downarrow$              & 3.32 & \rd{2.65} & 2.81 & \nd{1.51} & 2.20 & 3.44 & 3.27 & 3.78 & 2.87 \\
    & Comp.Ratio[$<\!5$cm\,\%]$\uparrow$ & 88.20 & 89.33 & \rd{89.10} & 93.41 & 90.14 & 85.64 & 85.47 & 85.30 & 88.43 \\
    \bottomrule
  \end{tabular}}
  \vspace{-1em}
  \label{tab:replica_rec}
\end{table}

\begin{table}[t]
  \caption{\textbf{Runtime and Memory Usage.} We comprehensively compared the runtime and memory usage on Replica~\cite{straub2019replica} Office 0. Our proposed CG-SLAM can perform more efficient tracking and mapping than existing works, actually reaching a real-time level. "-" indicates unavailable results in related works.}
  \vspace{-0.8em}
  \centering
  \footnotesize
  \setlength{\tabcolsep}{3pt}
  \renewcommand{\arraystretch}{1.0}
  \resizebox{0.8\linewidth}{!}{
  \begin{tabular}{@{\hspace*{1em}}lcccccc}
    \toprule
    \multirow{2}{*}{Method} 
    & \tt{Tracking} & \tt{Mapping} & \tt{Mapping} & \tt{System} & \tt{Decoder} & \tt{Scene}\\
    & \tt{[ms$\times$ it]$\downarrow$} & \tt{[ms$\times$ it]$\downarrow$} & \tt{Interval} & \tt{FPS$\uparrow$} & \tt{Param$\downarrow$} & \tt{Embeeding$\downarrow$} \\
    \midrule
    Vox-Fusion    &  23.61~$\times$~30  & 86.55~$\times$~10  &  10 & 1.1  &  0.98 MB  &  0.162 MB \\
    NICE-SLAM    & 6.19 $\times$ 10 & 91.59 $\times$ 60  &  5 & 0.98 &  0.43 MB  &  89.56 MB \\
    Co-SLAM      & 4.45 $\times$ 10  &  10.9 $\times$ 10 & 5 & \nd{14.2}  &  6.37 MB  &  - \\    
    Point-SLAM   & 6.14 $\times$ 40  & 22.25$\times$ 300 &  5 & 0.48 &  0.54 MB  &  28.11 MB \\
    GS-SLAM   & 11.9 $\times$ 10  & 12.8$\times$ 100 &  - & 8.34 &  -  &  - \\
    SplaTAM   & 41.7 $\times$ 40 & 50.1 $\times$ 60 &  1 & 0.21 &  -  &  - \\
    \midrule
    \textbf{Ours}     &  7.89 $\times$ 15 &  12.2 $\times$  60 & 30  & \rd{8.5}  & - &  231.66 MB \\
    \textbf{Ours-light}     &  3.80 $\times$  15 &  3.70 $\times$  60 & 30  & \fs{15.4}  &  - & 56.50 MB  \\
    \bottomrule
  \end{tabular}}
  \label{tab:Runtime And Memory Usage Analysis}
\end{table}
% Runtime Analysis on Replica Table.

% \begin{table}[t]
%   \caption{\textbf{Tracking Results across all versions.} (ATE RMSE [cm]$\downarrow$). Our lightweight version maintains competitive tracking performance on Replica~\cite{straub2019replica} scenes.}
%   \centering
%   \footnotesize
%   \setlength{\tabcolsep}{3pt}
%   \renewcommand{\arraystretch}{1.0}
%   \resizebox{0.57\linewidth}{!}{
%   \begin{tabular}{@{\hspace*{1em}}lccccccccc}
%     \toprule
%     & \multicolumn{1}{c}{\makecell{\tt{rm-0}}} & \multicolumn{1}{c}{\makecell{\tt{rm-1}}} &  \multicolumn{1}{c}{\makecell{\tt{rm-2}}} & \multicolumn{1}{c}{\makecell{\tt{off-0}}} & \multicolumn{1}{c}{\makecell{\tt{off-1}}} & \multicolumn{1}{c}{\makecell{\tt{off-2}}}& \multicolumn{1}{c}{\makecell{\tt{off-3}}} & \multicolumn{1}{c}{\makecell{\tt{off-4}}} & Avg. \\
%     \midrule
%     \textbf{Ours-full}   & \fs{0.29} & \fs{0.27} & \fs{0.25} & \nd{0.33} & \fs{0.14} & \fs{0.28} & \fs{0.31} & \fs{0.29} & \fs{0.27} \\
%     \textbf{Ours-light}   & \nd{0.44} & \nd{0.40} & \nd{0.34} & \fs{0.31} & \nd{0.30} & \nd{0.43} & \nd{0.48} & \nd{0.63} & \nd{0.42} \\
%     \bottomrule
%   \end{tabular}}
%   \label{tab:ablation-lighweight}
% \end{table}

\subsection{Runtime and Memory Analysis}
\label{4-3}
We evaluate the runtime and memory footprint of our system compared to other works in \cref{tab:Runtime And Memory Usage Analysis}. We reported the tracking and mapping efficiency in terms of per-iteration time consumption and the total number of optimization iterations. 
With the support of the GPU-accelerated rasterizer, our system can operate at around 8.5Hz. Meanwhile, our carefully designed pipeline allows this system to expand to a lightweight version and to work with half-resolution images. 
For tracking, this lightweight version can work twice as fast as the original one, at the cost of a slight decrease in accuracy, as shown in \cref{tab:Replica_tracking,tab:TUM-RGBD_tracking,tab:ScanNet_tracking}. For mapping, our lightweight version demonstrates similar superiority in \cref{tab:replica_rec}.
It can be clearly seen that both our versions achieve better performance than Co-SLAM~\cite{co-slam}. Further efficiency analysis on TUM-RGBD~\cite{sturm2012benchmark} and ScanNet~\cite{dai2017scannet} is shown in supplementary. Also, our customized Gaussian field allows us to outperform the concurrent Gaussian-based works with less computational burden.

% we also proposed a lightweight version that worked with half-resolution images. 
However, as a non-MLP scene representation, the 3D Gaussian field inevitably requires much memory consumption to store different properties. This reason results in a considerable memory footprint in the Gaussian-based SLAM system, which is a common limitation in other Gaussian-based research topics.

\subsection{Ablation Study}
\label{4-5}
To verify the rationality of our designs, we investigate the effectiveness of the anisotropy regularization, alignment and variance losses, and uncertainty model.

\begin{figure*}[!t]
  \centering
  \includegraphics[width=0.75\linewidth]{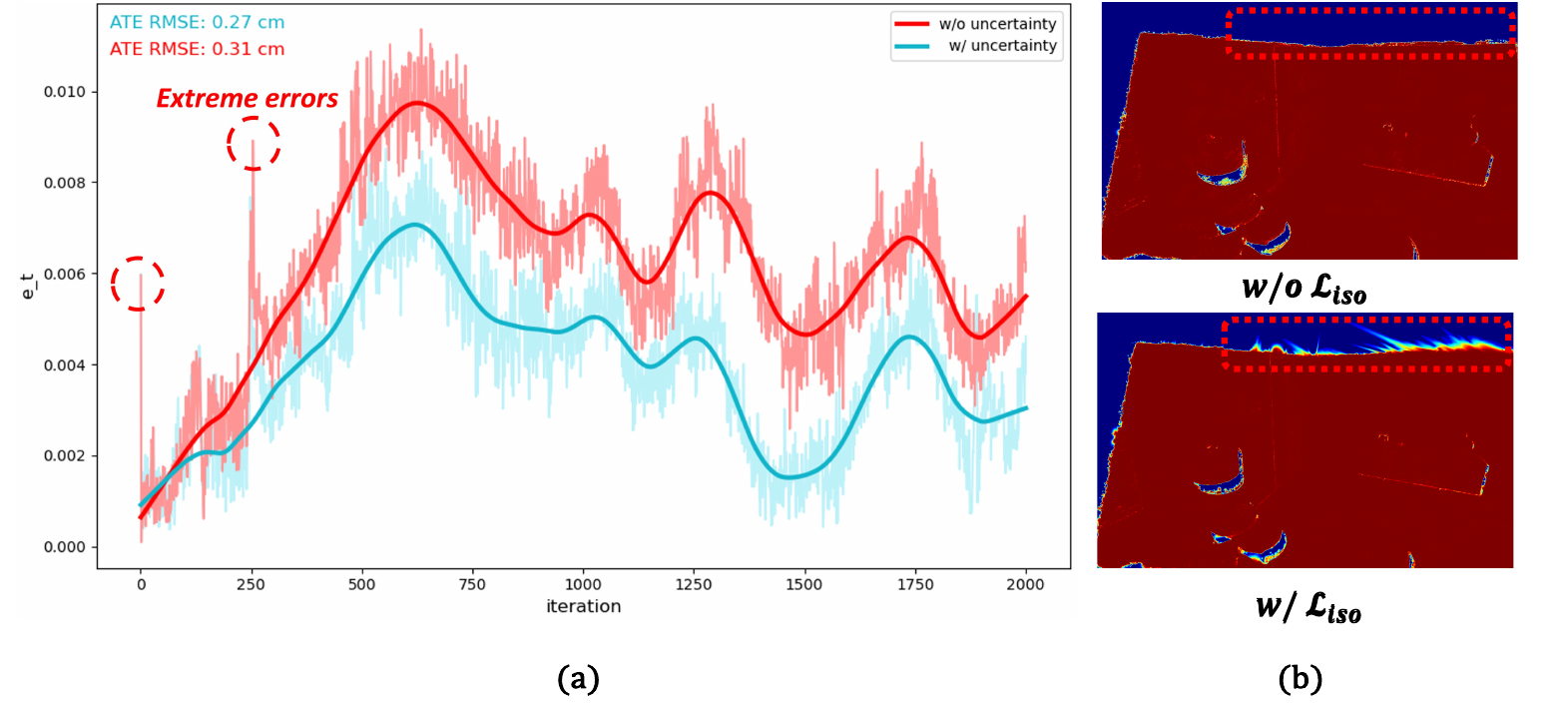}
  \caption{\textbf{Uncertainty Model Ablation and Anisotropy Interference.} (a) Uncertainty Model Ablation. This plot illustrates that the uncertainty model significantly helps improve tracking accuracy while avoiding some extreme errors. (b) Anisotropy Interference. It can be clearly seen that in the case of w/o $\mathcal{L}_{iso}$, serious arrow-shaped artifacts occur on the edges of the image. }
  \vspace{-0.8em}
  \label{fig:ablation_all}
\end{figure*}

\textbf{Effectiveness of Isotropy Loss.}
Ablation results in \cref{tab:ablation_isotropy} reveal how the anisotropy regularization term affects tracking metrics. We can notice that, in the absence of this term, sequential tracking will be limited by poor consistency, causing performance degradation. To more intuitively illustrate this phenomenon, we display opacity maps with and without anisotropy regularization in \cref{fig:ablation_all}~(b), and apparent arrow-shaped artifacts occur in the latter. Both quantitative and qualitative results support this design in our system.

\textbf{Effectiveness of Alignment and Variance Losses.}
We expect primitives that are closer to object surfaces in a 3D Gaussian field, whether for novel view synthesis or the SLAM task. To explain the contribution of our alignment and variance losses to the 3D Gaussian field, we add the $Chamfer~Distance$ as a new metric to quantify the distance between Gaussian primitives and surfaces. As shown in \cref{tab:ablation_alignment&variance}, these two loss functions effectively reduce the $Chamfer~Distance$ and further have a positive impact on the tracking metric.

\begin{table}[t]
  \caption{\textbf{Isotropy Loss Ablation Results(ATE RMSE [cm] $\downarrow$).} The experimental results demonstrate the effectiveness of our anisotropy regularization term. "-" indicates a failure situation.}
  \vspace{-1em}
  \centering
  \footnotesize
  \setlength{\tabcolsep}{3pt}
  \renewcommand{\arraystretch}{1.0}
  \resizebox{0.7\linewidth}{!}{
  \begin{tabular}{@{\hspace*{1em}}lccccccccc}
    \toprule{Setting}
    &\tt{rm-0} & \tt{rm-1} &  \tt{rm-2} & \tt{off-0} & \tt{off-1} & \tt{off-2}& \tt{off-3} & \tt{off-4} & Avg. \\
    \midrule
    \textbf{w/o $\mathcal{L}_{iso}$}   & \nd{0.32}    & \nd{0.31}    & \nd{0.54}    & \nd{0.36}    & \nd{0.24}    & \nd{0.31}    & \nd{0.72}    & -    & -    \\
    \textbf{Ours}   & \fs{0.29} & \fs{0.27} & \fs{0.25} & \fs{0.33} & \fs{0.14} & \fs{0.28} & \fs{0.31} & \fs{0.29} & \fs{0.27} \\
    \bottomrule
  \end{tabular}}
  \label{tab:ablation_isotropy}
\end{table}

\begin{table}[!t]
  \caption{\textbf{Alignment \& Variance Loss Ablation Results.} We analyze the effectiveness of alignment and variance losses using tracking and reconstruction metrics. We show average results from 8 Replica~\cite{straub2019replica} scenes in this table.}
  \vspace{-0.8em}
  \centering
  \footnotesize
  \setlength{\tabcolsep}{3pt}
  \renewcommand{\arraystretch}{1.0}
  \resizebox{0.7\linewidth}{!}{
  \begin{tabular}{lccccccccccccccccccc}
    \toprule
    Metric & \multicolumn{1}{c}{\makecell{\tt{w/o $\mathcal{L}_{align}$}}} & \multicolumn{1}{c}{\makecell{\tt{w/o $\mathcal{L}_{Var}$}}} &  \multicolumn{1}{c}{\makecell{\tt{w/o $\mathcal{L}_{align}$\&$\mathcal{L}_{var}$}}} & \multicolumn{1}{c}{\makecell{\tt{\textbf{Ours-full}}}} \\
    \midrule    
    RMSE.[cm]$\downarrow$  & \nd{0.28} & \rd{0.30} & 0.33 & \fs{0.26} \\
    % Mean[cm]$\downarrow$   & \nd{0.25} & \nd{0.25} & \nd{0.25} & \fs{0.24} \\
    % Median[cm]$\downarrow$ & \nd{0.24} & \nd{0.24} & \nd{0.24} & \fs{0.23} \\
    Chamfer dis.[cm]$\downarrow$ & \rd{4.74} & \nd{4.57} & 4.79 & \fs{3.85} \\
    \bottomrule
  \end{tabular}}
  \vspace{-1.4em}
  \label{tab:ablation_alignment&variance}
\end{table}

\textbf{Effectiveness of Uncertainty Model.} As shown in \cref{fig:ablation_all}~(a), we have elaborately plotted curves of tracking errors in the Replica~\cite{straub2019replica} Room 1 in two cases: w/ and w/o the uncertainty model. This visualization illustrates the rationality of our depth uncertainty model and the importance of geometric cues for the 3D Gaussian field. Specifically, removing Gaussian primitives with large uncertainty can effectively help improve localization accuracy while avoiding some extreme errors to reduce localization variance.

\section{Conclusion}
We have proposed CG-SLAM, a dense RGB-D SLAM based on a consistent and uncertainty-aware 3D Gaussian field. Our targeted loss functions strengthen the 3D Gaussian field in terms of consistency and stability. The uncertainty model further distills highly informative primitives in this field to reduce interference from outliers. Besides, a customized GPU-accelerated rasterization pipeline enables our system to achieve state-of-the-art accuracy and efficiency in various scenes. Through extensive experiments, it can be concluded that our method outperforms previous works regarding tracking, mapping, and efficiency. 

\textbf{Limitations.} 
Considerable memory usage is one limitation of the Gaussian-based system. We expect that a more compact Gaussian field can be adopted in the SLAM task. In addition, the Gaussian-based method has a weak prediction ability for unobserved areas. Moreover, our system is not capable of handling dynamic objects in the environment. We believe it is a very interesting direction for future work.

\section*{Acknowledgement} 
This work was partially supported by the NSFC (No. 62102356).

% \clearpage  % TODO REVIEW/FINAL: This \clearpage needs to be removed from both review and camera-ready versions.

\bibliographystyle{splncs04}
\bibliography{main}

\begin{thebibliography}{10}
\providecommand{\url}[1]{\texttt{#1}}
\providecommand{\urlprefix}{URL }
\providecommand{\doi}[1]{https://doi.org/#1}

\bibitem{gaussianshell}
Abdal, R., Yifan, W., Shi, Z., Xu, Y., Po, R., Kuang, Z., Chen, Q., Yeung, D.Y., Wetzstein, G.: Gaussian shell maps for efficient 3d human generation (2023)

\bibitem{arandjelovic2016netvlad}
Arandjelovic, R., Gronat, P., Torii, A., Pajdla, T., Sivic, J.: Netvlad: Cnn architecture for weakly supervised place recognition. In: Proceedings of the IEEE conference on computer vision and pattern recognition. pp. 5297--5307 (2016)

\bibitem{mipnerf}
Barron, J.T., Mildenhall, B., Tancik, M., Hedman, P., Martin-Brualla, R., Srinivasan, P.P.: Mip-nerf: A multiscale representation for anti-aliasing neural radiance fields (2021)

\bibitem{mipnerf360}
Barron, J.T., Mildenhall, B., Verbin, D., Srinivasan, P.P., Hedman, P.: Mip-nerf 360: Unbounded anti-aliased neural radiance fields. CVPR  (2022)

\bibitem{neusg}
Chen, H., Li, C., Lee, G.H.: Neusg: Neural implicit surface reconstruction with 3d gaussian splatting guidance (2023)

\bibitem{ChenBI13}
Chen, J., Bautembach, D., Izadi, S.: Scalable real-time volumetric surface reconstruction. {ACM} Trans. Graph.  \textbf{32}(4),  113:1--113:16 (2013). \doi{10.1145/2461912.2461940}, \url{https://doi.org/10.1145/2461912.2461940}

\bibitem{GaussianEditor}
Chen, Y., Chen, Z., Zhang, C., Wang, F., Yang, X., Wang, Y., Cai, Z., Yang, L., Liu, H., Lin, G.: Gaussianeditor: Swift and controllable 3d editing with gaussian splatting (2023)

\bibitem{orbeez}
Chung, C., Tseng, Y., Hsu, Y., Shi, X.Q., Hua, Y., Yeh, J., Chen, W., Chen, Y., Hsu, W.H.: Orbeez-slam: {A} real-time monocular visual {SLAM} with {ORB} features and nerf-realized mapping. In: {IEEE} International Conference on Robotics and Automation, {ICRA} 2023, London, UK, May 29 - June 2, 2023. pp. 9400--9406. {IEEE} (2023). \doi{10.1109/ICRA48891.2023.10160950}, \url{https://doi.org/10.1109/ICRA48891.2023.10160950}

\bibitem{tsdf-fusion}
Curless, B., Levoy, M.: A volumetric method for building complex models from range images. In: Proceedings of the 23rd annual conference on Computer graphics and interactive techniques. pp. 303--312 (1996)

\bibitem{dai2017scannet}
Dai, A., Chang, A.X., Savva, M., Halber, M., Funkhouser, T., Nie{\ss}ner, M.: Scannet: Richly-annotated 3d reconstructions of indoor scenes. In: Proceedings of the IEEE conference on computer vision and pattern recognition. pp. 5828--5839 (2017)

\bibitem{mdsplatting}
Duisterhof, B.P., Mandi, Z., Yao, Y., Liu, J.W., Shou, M.Z., Song, S., Ichnowski, J.: Md-splatting: Learning metric deformation from 4d gaussians in highly deformable scenes (2023)

\bibitem{GaussianEditor_Fang}
Fang, J., Wang, J., Zhang, X., Xie, L., Tian, Q.: Gaussianeditor: Editing 3d gaussians delicately with text instructions. arXiv preprint arXiv:2311.16037  (2023)

\bibitem{TiNeuVox}
Fang, J., Yi, T., Wang, X., Xie, L., Zhang, X., Liu, W., Nie\ss{}ner, M., Tian, Q.: Fast dynamic radiance fields with time-aware neural voxels. In: SIGGRAPH Asia 2022 Conference Papers (2022)

\bibitem{plenoxels}
{Fridovich-Keil and Yu}, Tancik, M., Chen, Q., Recht, B., Kanazawa, A.: Plenoxels: Radiance fields without neural networks. In: CVPR (2022)

\bibitem{instant_nvr}
Geng, C., Peng, S., Xu, Z., Bao, H., Zhou, X.: Learning neural volumetric representations of dynamic humans in minutes. In: CVPR (2023)

\bibitem{sugar}
Gu{\'e}don, A., Lepetit, V.: Sugar: Surface-aligned gaussian splatting for efficient 3d mesh reconstruction and high-quality mesh rendering. arXiv preprint arXiv:2311.12775  (2023)

\bibitem{neo360}
Irshad, M.Z., Zakharov, S., Liu, K., Guizilini, V., Kollar, T., Gaidon, A., Kira, Z., Ambrus, R.: Neo 360: Neural fields for sparse view synthesis of outdoor scenes (2023), \url{https://arxiv.org/abs/2308.12967}

\bibitem{splatarmor}
Jena, R., Iyer, G.S., Choudhary, S., Smith, B., Chaudhari, P., Gee, J.: Splatarmor: Articulated gaussian splatting for animatable humans from monocular rgb videos. arXiv preprint arXiv:2311.10812  (2023)

\bibitem{nerfvins}
Katragadda, S., Lee, W., Peng, Y., Geneva, P., Chen, C., Guo, C., Li, M., Huang, G.: Nerf-vins: {A} real-time neural radiance field map-based visual-inertial navigation system. CoRR  \textbf{abs/2309.09295} (2023). \doi{10.48550/ARXIV.2309.09295}, \url{https://doi.org/10.48550/arXiv.2309.09295}

\bibitem{splatam}
Keetha, N., Karhade, J., Jatavallabhula, K.M., Yang, G., Scherer, S., Ramanan, D., Luiten, J.: Splatam: Splat, track \& map 3d gaussians for dense rgb-d slam. arXiv preprint arXiv:2312.02126  (2023)

\bibitem{point-fusion}
Keller, M., Lefloch, D., Lambers, M., Izadi, S., Weyrich, T., Kolb, A.: Real-time 3d reconstruction in dynamic scenes using point-based fusion. In: 2013 International Conference on 3D Vision, 3DV 2013, Seattle, Washington, USA, June 29 - July 1, 2013. pp.~1--8. {IEEE} Computer Society (2013). \doi{10.1109/3DV.2013.9}, \url{https://doi.org/10.1109/3DV.2013.9}

\bibitem{3Dgs}
Kerbl, B., Kopanas, G., Leimk{\"u}hler, T., Drettakis, G.: 3d gaussian splatting for real-time radiance field rendering. ACM Transactions on Graphics  \textbf{42}(4) (2023)

\bibitem{DVOslam}
Kerl, C., Sturm, J., Cremers, D.: Dense visual slam for rgb-d cameras. In: 2013 IEEE/RSJ International Conference on Intelligent Robots and Systems. pp. 2100--2106. IEEE (2013)

\bibitem{tandem}
Koestler, L., Yang, N., Zeller, N., Cremers, D.: Tandem: Tracking and dense mapping in real-time using deep multi-view stereo. In: Conference on Robot Learning (CoRL) (2021)

\bibitem{barf}
Lin, C.H., Ma, W.C., Torralba, A., Lucey, S.: Barf: Bundle-adjusting neural radiance fields. In: IEEE International Conference on Computer Vision ({ICCV}) (2021)

\bibitem{nerf}
Mildenhall, B., Srinivasan, P.P., Tancik, M., Barron, J.T., Ramamoorthi, R., Ng, R.: {NeRF}: Representing scenes as neural radiance fields for view synthesis. In: The European Conference on Computer Vision (ECCV) (2020)

\bibitem{instant}
M\"uller, T., Evans, A., Schied, C., Keller, A.: Instant neural graphics primitives with a multiresolution hash encoding. ACM Trans. Graph.  \textbf{41}(4),  102:1--102:15 (Jul 2022). \doi{10.1145/3528223.3530127}, \url{https://doi.org/10.1145/3528223.3530127}

\bibitem{KinectFusion}
Newcombe, R.A., Izadi, S., Hilliges, O., Molyneaux, D., Kim, D., Davison, A.J., Kohli, P., Shotton, J., Hodges, S., Fitzgibbon, A.W.: Kinectfusion: Real-time dense surface mapping and tracking. In: 10th {IEEE} International Symposium on Mixed and Augmented Reality, {ISMAR} 2011, Basel, Switzerland, October 26-29, 2011. pp. 127--136. {IEEE} Computer Society (2011). \doi{10.1109/ISMAR.2011.6092378}, \url{https://doi.org/10.1109/ISMAR.2011.6092378}

\bibitem{DTAM}
Newcombe, R.A., Lovegrove, S., Davison, A.J.: {DTAM:} dense tracking and mapping in real-time. In: Metaxas, D.N., Quan, L., Sanfeliu, A., Gool, L.V. (eds.) {IEEE} International Conference on Computer Vision, {ICCV} 2011, Barcelona, Spain, November 6-13, 2011. pp. 2320--2327. {IEEE} Computer Society (2011). \doi{10.1109/ICCV.2011.6126513}, \url{https://doi.org/10.1109/ICCV.2011.6126513}

\bibitem{MatthiasVoxel}
Nie{\ss}ner, M., Zollh{\"{o}}fer, M., Izadi, S., Stamminger, M.: Real-time 3d reconstruction at scale using voxel hashing. {ACM} Trans. Graph.  \textbf{32}(6),  169:1--169:11 (2013). \doi{10.1145/2508363.2508374}, \url{https://doi.org/10.1145/2508363.2508374}

\bibitem{voxblox}
Oleynikova, H., Taylor, Z., Fehr, M., Siegwart, R., Nieto, J.I.: Voxblox: Incremental 3d euclidean signed distance fields for on-board {MAV} planning. In: 2017 {IEEE/RSJ} International Conference on Intelligent Robots and Systems, {IROS} 2017, Vancouver, BC, Canada, September 24-28, 2017. pp. 1366--1373. {IEEE} (2017). \doi{10.1109/IROS.2017.8202315}, \url{https://doi.org/10.1109/IROS.2017.8202315}

\bibitem{ash}
Pang, H., Zhu, H., Kortylewski, A., Theobalt, C., Habermann, M.: Ash: Animatable gaussian splats for efficient and photoreal human rendering  (2023)

\bibitem{nerfslam}
Rosinol, A., Leonard, J.J., Carlone, L.: Nerf-slam: Real-time dense monocular {SLAM} with neural radiance fields. In: {IROS}. pp. 3437--3444 (2023). \doi{10.1109/IROS55552.2023.10341922}, \url{https://doi.org/10.1109/IROS55552.2023.10341922}

\bibitem{point-slam}
Sandstr{\"o}m, E., Li, Y., Van~Gool, L., Oswald, M.R.: Point-slam: Dense neural point cloud-based slam. In: Proceedings of the IEEE/CVF International Conference on Computer Vision. pp. 18433--18444 (2023)

\bibitem{sandstrom2023uncle}
Sandstr{\"o}m, E., Ta, K., Van~Gool, L., Oswald, M.R.: Uncle-slam: Uncertainty learning for dense neural slam. arXiv preprint arXiv:2306.11048  (2023)

\bibitem{badslam}
Schops, T., Sattler, T., Pollefeys, M.: Bad slam: Bundle adjusted direct rgb-d slam. In: Proceedings of the IEEE/CVF Conference on Computer Vision and Pattern Recognition (CVPR) (June 2019)

\bibitem{Graf}
Schwarz, K., Liao, Y., Niemeyer, M., Geiger, A.: Graf: Generative radiance fields for 3d-aware image synthesis. In: Advances in Neural Information Processing Systems (NeurIPS) (2020)

\bibitem{tensor4d}
Shao, R., Zheng, Z., Tu, H., Liu, B., Zhang, H., Liu, Y.: Tensor4d: Efficient neural 4d decomposition for high-fidelity dynamic reconstruction and rendering. In: Proceedings of the IEEE Conference on Computer Vision and Pattern Recognition (2023)

\bibitem{Franklarge}
Steinbr{\"{u}}cker, F., Kerl, C., Cremers, D.: Large-scale multi-resolution surface reconstruction from {RGB-D} sequences. In: {IEEE} International Conference on Computer Vision, {ICCV} 2013, Sydney, Australia, December 1-8, 2013. pp. 3264--3271. {IEEE} Computer Society (2013). \doi{10.1109/ICCV.2013.405}, \url{https://doi.org/10.1109/ICCV.2013.405}

\bibitem{straub2019replica}
Straub, J., Whelan, T., Ma, L., Chen, Y., Wijmans, E., Green, S., Engel, J.J., Mur-Artal, R., Ren, C., Verma, S., Clarkson, A., Yan, M., Budge, B., Yan, Y., Pan, X., Yon, J., Zou, Y., Leon, K., Carter, N., Briales, J., Gillingham, T., Mueggler, E., Pesqueira, L., Savva, M., Batra, D., Strasdat, H.M., Nardi, R.D., Goesele, M., Lovegrove, S., Newcombe, R.: The replica dataset: A digital replica of indoor spaces (2019)

\bibitem{sturm2012benchmark}
Sturm, J., Engelhard, N., Endres, F., Burgard, W., Cremers, D.: A benchmark for the evaluation of rgb-d slam systems. In: 2012 IEEE/RSJ international conference on intelligent robots and systems. pp. 573--580. IEEE (2012)

\bibitem{imap}
Sucar, E., Liu, S., Ortiz, J., Davison, A.J.: imap: Implicit mapping and positioning in real-time. In: Proceedings of the IEEE/CVF International Conference on Computer Vision. pp. 6229--6238 (2021)

\bibitem{cnnslam}
Tateno, K., Tombari, F., Laina, I., Navab, N.: {CNN-SLAM:} real-time dense monocular {SLAM} with learned depth prediction. In: 2017 {IEEE} Conference on Computer Vision and Pattern Recognition, {CVPR} 2017, Honolulu, HI, USA, July 21-26, 2017. pp. 6565--6574. {IEEE} Computer Society (2017). \doi{10.1109/CVPR.2017.695}, \url{https://doi.org/10.1109/CVPR.2017.695}

\bibitem{droidslam}
Teed, Z., Deng, J.: {DROID-SLAM:} deep visual {SLAM} for monocular, stereo, and {RGB-D} cameras. In: Ranzato, M., Beygelzimer, A., Dauphin, Y.N., Liang, P., Vaughan, J.W. (eds.) Advances in Neural Information Processing Systems 34: Annual Conference on Neural Information Processing Systems 2021, NeurIPS 2021, December 6-14, 2021, virtual. pp. 16558--16569 (2021), \url{https://proceedings.neurips.cc/paper/2021/hash/89fcd07f20b6785b92134bd6c1d0fa42-Abstract.html}

\bibitem{scenerflow}
Tretschk, E., Golyanik, V., Zollh\"{o}fer, M., Bozic, A., Lassner, C., Theobalt, C.: Scenerflow: Time-consistent reconstruction of general dynamic scenes. In: International Conference on 3D Vision (3DV) (2024)

\bibitem{efficientoctree}
Vespa, E., Nikolov, N., Grimm, M., Nardi, L., Kelly, P.H.J., Leutenegger, S.: Efficient octree-based volumetric {SLAM} supporting signed-distance and occupancy mapping. {IEEE} Robotics Autom. Lett.  \textbf{3}(2),  1144--1151 (2018). \doi{10.1109/LRA.2018.2792537}, \url{https://doi.org/10.1109/LRA.2018.2792537}

\bibitem{co-slam}
Wang, H., Wang, J., Agapito, L.: Co-slam: Joint coordinate and sparse parametric encodings for neural real-time slam. In: Proceedings of the IEEE/CVF Conference on Computer Vision and Pattern Recognition. pp. 13293--13302 (2023)

\bibitem{rodin}
Wang, T., Zhang, B., Zhang, T., Gu, S., Bao, J., Baltrusaitis, T., Shen, J., Chen, D., Wen, F., Chen, Q., et~al.: Rodin: A generative model for sculpting 3d digital avatars using diffusion. arXiv preprint arXiv:2212.06135  (2022)

\bibitem{elastic}
Whelan, T., Salas{-}Moreno, R.F., Glocker, B., Davison, A.J., Leutenegger, S.: Elasticfusion: Real-time dense {SLAM} and light source estimation. Int. J. Robotics Res.  \textbf{35}(14),  1697--1716 (2016). \doi{10.1177/0278364916669237}, \url{https://doi.org/10.1177/0278364916669237}

\bibitem{pointnerf}
Xu, Q., Xu, Z., Philip, J., Bi, S., Shu, Z., Sunkavalli, K., Neumann, U.: Point-nerf: Point-based neural radiance fields. arXiv preprint arXiv:2201.08845  (2022)

\bibitem{gsslam}
Yan, C., Qu, D., Wang, D., Xu, D., Wang, Z., Zhao, B., Li, X.: Gs-slam: Dense visual slam with 3d gaussian splatting (2024)

\bibitem{D3VO}
Yang, N., von Stumberg, L., Wang, R., Cremers, D.: {D3VO:} deep depth, deep pose and deep uncertainty for monocular visual odometry. In: 2020 {IEEE/CVF} Conference on Computer Vision and Pattern Recognition, {CVPR} 2020, Seattle, WA, USA, June 13-19, 2020. pp. 1278--1289. Computer Vision Foundation / {IEEE} (2020). \doi{10.1109/CVPR42600.2020.00136}, \url{https://openaccess.thecvf.com/content\_CVPR\_2020/html/Yang\_D3VO\_Deep\_Depth\_Deep\_Pose\_and\_Deep\_Uncertainty\_for\_Monocular\_CVPR\_2020\_paper.html}

\bibitem{vox-fusion}
Yang, X., Li, H., Zhai, H., Ming, Y., Liu, Y., Zhang, G.: Vox-fusion: Dense tracking and mapping with voxel-based neural implicit representation. In: 2022 IEEE International Symposium on Mixed and Augmented Reality (ISMAR). pp. 499--507. IEEE (2022)

\bibitem{gau-grouping}
Ye, M., Danelljan, M., Yu, F., Ke, L.: Gaussian grouping: Segment and edit anything in 3d scenes. arXiv preprint arXiv:2312.00732  (2023)

\bibitem{pixelnerf}
Yu, A., Ye, V., Tancik, M., Kanazawa, A.: {pixelNeRF}: Neural radiance fields from one or few images. In: CVPR (2021)

\bibitem{gavatar}
Yuan, Y., Li, X., Huang, Y., De~Mello, S., Nagano, K., Kautz, J., Iqbal, U.: Gavatar: Animatable 3d gaussian avatars with implicit mesh learning. arXiv preprint arXiv:2312.11461  (2023)

\bibitem{featuredis}
Zhou, S., Chang, H., Jiang, S., Fan, Z., Zhu, Z., Xu, D., Chari, P., You, S., Wang, Z., Kadambi, A.: Feature 3dgs: Supercharging 3d gaussian splatting to enable distilled feature fields. arXiv preprint arXiv:2312.03203  (2023)

\bibitem{nice-slam}
Zhu, Z., Peng, S., Larsson, V., Xu, W., Bao, H., Cui, Z., Oswald, M.R., Pollefeys, M.: Nice-slam: Neural implicit scalable encoding for slam. In: Proceedings of the IEEE/CVF Conference on Computer Vision and Pattern Recognition. pp. 12786--12796 (2022)

\bibitem{Drivable3D}
Zielonka, W., Bagautdinov, T., Saito, S., Zollhöfer, M., Thies, J., Romero, J.: Drivable 3d gaussian avatars  (2023)

\bibitem{zwicker2001ewa}
Zwicker, M., Pfister, H., Van~Baar, J., Gross, M.: Ewa volume splatting. In: Proceedings Visualization, 2001. VIS'01. pp. 29--538. IEEE (2001)

\end{thebibliography}
\end{document}